\documentclass{article}

\usepackage[final, nonatbib]{drlw_neurips_2019}

\usepackage[utf8]{inputenc} 
\usepackage[T1]{fontenc}    
\usepackage{hyperref}       
\usepackage{url}            
\usepackage{booktabs}       
\usepackage{amsfonts}       
\usepackage{nicefrac}       
\usepackage{microtype}      

\usepackage{mathtools}
\usepackage{multirow}
\usepackage{bbold}

\usepackage[backend=biber,style=apa, backref=true, sortcites=true,sorting=nyt, uniquename=false]{biblatex}
\DeclareLanguageMapping{american}{american-apa}
\usepackage[american]{babel}
\usepackage{hyperref}
\usepackage{arydshln}

\title{Learning Sparse Representations Incrementally in Deep Reinforcement Learning}

\author{%
  J. Fernando Hernandez-Garcia \\
  Department of Computing Science \\
  University of Alberta \\
  Edmonton, AB, T6G 2E8 \\
  \texttt{jfhernan@ualberta.ca} \\
   \And
  Richard S. Sutton \\
  Department of Computing Science \\
  University of Alberta \\
  Edmonton, AB, T6G 2E8 \\
  \texttt{rsutton@ualberta.ca}
}

\begin{document}

\maketitle

\begin{abstract}
Sparse representations have been shown to be useful in deep reinforcement learning for mitigating catastrophic interference and improving the performance of agents in terms of cumulative reward.
Previous results were based on a two step process were the representation was learned offline and the  action-value function was learned online afterwards.
In this paper, we investigate if it is possible to learn a sparse representation and the action-value function simultaneously and incrementally.
We investigate this question by employing several regularization techniques and observing how they affect sparsity of the representation learned by a DQN agent in two different benchmark domains.
Our results show  that with appropriate regularization it is possible to increase the sparsity of the representations learned by DQN agents.
Moreover, we found that learning sparse representations also resulted in improved performance in terms of cumulative reward.
Finally, we found that the performance of the agents that learned a sparse representation was more robust to the size of the experience replay buffer. 
This last finding supports the long standing hypothesis that the overlap in representations learned by deep neural networks is the leading cause of catastrophic interference.
%
\end{abstract}

In reinforcement learning (RL) it is of interest to learn incrementally as agents interact with their environment.
Moreover, we also care for algorithms that are scalable to high-dimensional environments because such are the type of environments we expect to encounter in the real world.
Since they are readily scalable to high-dimensional data, deep neural networks (DNNs) have become popular for learning action-value functions resulting in many successes in complex, high-dimensional environments \parencite{Mnih2015HumanLevel, Silver2017Mastering, Jaderberg2016Unreal}.
However, DNNs have proven to be difficult to train incrementally.
To address this issue, techniques such as target networks, inspired by batch methods such as Fitted Q-iteration \parencite{Riedmiller2005FittedQ}, and experience replay \parencite{Lin1992RLforRobots} are often employed to facilitate training. 

We argue that one of the reasons that DNNs are difficult to train in RL is because of \textit{catastrophic interference}---the inability of a neural network to remember old information when learning from new observations.
To support this claim it is key to emphasize that both target networks and experience replay can be viewed as techniques for reducing the amount of interference inherent to the \textit{distributed representations}---the share set of features---found in DNNs.
Target networks prevent abrupt changes to the weights corresponding to one region of the state space from arbitrarily influencing the values corresponding to other regions, undoing the learning done so far.
The experience replay buffer prevents knowledge of previous observations from being forgotten by storing a buffer of observations and interleaving training on old and new observations.
Thus, both of these techniques mitigate the catastrophic interference experienced by DNNs.
In fact, they can be categorized as two examples from the five main approaches for mitigating catastrophic interference outlined by \citeauthor{Kemker2018MeasuringCF} (\citeyear{Kemker2018MeasuringCF}).
The target network can be viewed as a simple form of dual-memory model where old representations are kept separate from new ones to prevent overriding old information, whereas experience replay is a rehearsal method that allows the network to retain its knowledge of old observations by constantly retraining on those observations.

Characterizing the difficulty of training DNNs for RL as an instance of catastrophic interference is useful because it allows us to consider a body of literature that has been developed for over 30 years--- since \citeauthor{McCloskey1989Catastrophic} (\citeyear{McCloskey1989Catastrophic}) and \citeauthor{Ratcliff1990Connectionist} (\citeyear{Ratcliff1990Connectionist}) first pointed out the occurrence of catastrophic interference.
The leading hypothesis is that the overlap in representations in DNNs are the main source of interference because modifying one representation given a particular observation inevitably affects the representations corresponding to other possible observations \parencite{French1991UsingSR}. 
To illustrate how to overcome catastrophic interference, consider the case of a lookup table representation with an entry for each different possible observation.
Lookup table representations, unlike the distributed representations learned in DNNs, do not share any features; hence, interference cannot possibly occur.
The downside is that lookup table representations do not scale well with the size of the observation space and learning does not generalize across observations.
Consequently, it is desirable to learn representations that can generalize across different observations, but that are local enough that changing the features for a given observation only affects the representation of other similar observations: a \textit{semi-distributed representation}.

One way to achieve semi-distributed representations is by using \textit{sparse representations} where only a small set of features is active (they are non-zero) for any given observation.
In reinforcement learning, sparse representations with hand-crafted features such as tile-coding, radial basis functions, and sparse distributed memory are commonly used for control, where the goal of an agent is to maximize the amount of reward that it receives \parencite{Sutton1988Learning, Sutton2018Book, BohdanaPrecup2004SDM}.
These types of representations facilitate incremental learning, but they are hard to scale to problems with high-dimensional observations.
Hence, a viable alternative is to train DNNs in such a way so that the learned representation is sparse.

The utility of sparse representations in deep RL has already been studied in the past with beneficial results for mitigating catastrophic interference and improving performance in terms of cumulative reward \parencite{Liu2018SRUtility, Lei2017LearningSR}. 
In both of these studies, training followed a two step process where the representation was learned first and the action-value function was learned afterwards.
In this paper, we extend the work by \citeauthor{Liu2018SRUtility} (\citeyear{Liu2018SRUtility}) by investigating if similar results can be found when the representation and the action-value function are learned simultaneously and incrementally.
We demonstrate empirically in two different domains that with appropriate regularization it is possible to learn a sparse representation incrementally in a DQN agent, and that learning such representation also results in improved performance.
Additionally, we show that the performance of DQN agents that learn a sparse representation is more robust to the size of the replay buffer, which provides evidence of the utility of sparse representations for overcoming catastrophic interference and complements the results from \citeauthor{Liu2018SRUtility} (\citeyear{Liu2018SRUtility}). 

\section{Background and Notation}

In reinforcement learning, the sequential decision-making problem is modeled using the Markov Decision Process formulation defined by the tuple $\langle \mathcal{S}, \mathcal{A}, \mathcal{R}, P, \gamma \rangle$.
In this formulation, an agent---which is both a learner and an actor---and an environment interact over a sequence of discrete time steps $t \geq 0$. 
At every time step the agent observes a state $S_t \!\in\! \mathcal{S}$, which encodes information about the environment.
Based on that information, the agent chooses and executes an action $A_t \!\in\! \mathcal{A}$. 
As a consequence of the executed action, the environment sends back to the agent a new state $S_{t+1}$ and a reward $R_{t+1}\! \in \!\mathcal{R} \! \subseteq \!\mathbb{R}$.
The reward and new state are modeled jointly by the transition dynamics probability function $P$, which defines the probability of observing $(S_{t+1}, R_{t+1})$ given $(S_t, A_t)$.

Actions are selected according to a policy $\pi(\cdot | S_{t})$---a probability distribution over the actions given the current state.
The goal of the agent is to maximize the expected sum of discounted rewards $\mathbb{E}_\pi \big\{ \sum_{i \geq t} \gamma^{i - t} R_{i+1} \big\}$, where the expectation is with respect to $\pi$ and $P$, and $\gamma$ is a discount factor in the interval $[0,1]$. 
In order to make informed decisions, agents often estimate the action-value function $q_\pi$ which maps states and actions to a value in $\mathbb{R}$ and is defined as
\begin{equation}
\label{eq:qpi}
    q_\pi (s,a) \overset{.}{=} \mathbb{E}_\pi \Big \{ \sum_{i \geq t} \gamma^{i-t} R_{i+1} \big | S_t = s, A_t = a \Big \}.
\end{equation}
Then, through the process of policy iteration \parencite{Sutton2018Book}, the agent can learn the optimal action-value function $q_*(s,a) \overset{.}{=}\max_\pi q_\pi (s,a)$ for all state-action pairs $(s,a)$.

The optimal action-value function obeys an important relationship called the \textit{Bellman equation}, which relates the action-value for a state-action pair at time $t$ to the action-values at the next time-step.
Using this identity, we can model stochastic approximation algorithms for estimating $q_*$. 
One of the most popular of these algorithms is the Q-Learning algorithm \parencite{Watkins1989QLearn} which iteratively computes estimates of the action-value function, $Q$, using the update rule:
\begin{equation}
\label{eq:qlearning}
    Q_{t+1}(S_t, A_t) \leftarrow Q_{t}(S_t, A_t) + \alpha [R_{t+1} + \gamma \max_a Q_t(S_{t+1},a) - Q_{t}(S_t, A_t) ], t \geq 0,
\end{equation}
where $Q_{t+1}(s,a) = Q_t(s,a)$ for all $(s,a) \neq (S_t,A_t)$ and $Q_0$ is initialized arbitrarily.

\subsection{Function Approximation}

The update rule in Equation (\ref{eq:qlearning}) can be easily implemented with a lookup table representation. 
However, we are interested in the case where the action-value is approximated as a function of a vector of parameters $\textbf{w} \in \mathbb{R}^d$, $d \in \mathbb{N}$, and a parameterized representation $\boldsymbol{\phi}_{\boldsymbol{\theta}}: \mathcal{S} \times \mathcal{A} \rightarrow \mathbb{R}^d$.
In such a case, we approximate $q_*$ as:
\begin{equation}
    \label{eq:approx_av}
    q_*(s,a) \approx \Hat{q}(s,a, \boldsymbol{\theta}, \textbf{w}) \overset{.}{=} \textbf{w}^\top \boldsymbol{\phi}_{\boldsymbol{\theta}}(s,a) , \ (s,a) \in \mathcal{S} \times \mathcal{A}.
\end{equation}
Specifically, we want $\boldsymbol{\phi}_{\boldsymbol{\theta}}$ to be a sparse representation with very few active (non-zero) features for any given state-action pair. 

In their paper, \citeauthor{Liu2018SRUtility} (\citeyear{Liu2018SRUtility}) used the last layer of a fully-connected neural network as the representation $\boldsymbol{\phi}_{\boldsymbol{\theta}}$ and learned the parameters $\boldsymbol{\theta}$ using stochastic gradient descent to minimize the Mean-Squared Temporal Difference Error of a fixed policy.
After learning the representation, they learned the weights $\textbf{w}$ using the semi-gradient version of the Sarsa(0) algorithm---an on-policy alternative to Q-Learning \parencite{Rummery1995Sarsa, Singh2000SarsaZ, Sutton1996Generalization, Sutton1998Book}.

In this work, we are interested in learning both the representation $\boldsymbol{\phi}_{\boldsymbol{\theta}}$ and the weight vector $\textbf{w}$ simultaneously. 
Hence, we will model both sets of parameters $\boldsymbol{\theta}$ and $\textbf{w}$ as part of a single feed-forward neural network and learn them using the well-known DQN architecture \parencite{Mnih2015HumanLevel, Mnih2013PlayingAtari}.
DQN seeks to minimize the loss function:
\begin{equation}
    \label{eq:dqn_loss}
    L(\boldsymbol{\theta}_t) \overset{.}{=} \mathbb{E}_\pi 
    \Big\{ \big( R_{t+1} + \gamma \max_a \hat{q}(S_{t+1}, a, \boldsymbol{\theta}^-_t) - \hat{q}(S_t, A_t, \boldsymbol{\theta}_t) \big)^2 \big| S_t \!=\!s, A_t \!=\! a  \Big\},
\end{equation}
where $\hat{q}$ is a neural network parameterized by $\boldsymbol{\theta}_t$---the \textit{policy network}---and $\boldsymbol{\theta}^-_t$ is a separate set of parameters---the \textit{target network}---that is updated every certain number of training steps by setting it equal to $\boldsymbol{\theta}_t$.
It is important to emphasize that $\boldsymbol{\theta}_t$ is used when selecting actions and is updated at every training step, whereas $\boldsymbol{\theta}^-_t$ is exclusively used to compute the loss function. 
To minimize the loss function we compute stochastic gradient descent updates on a mini-batch of transitions sampled from the experience replay buffer, which stores transitions of the form $(S_k, A_k, R_k, S_{k+1})$ for $k \leq t\!-\!1$.

\section{Regularization Techniques}

In order to learn a sparse representation while training the DQN architecture, we will employ similar regularization techniques as in \citeauthor{Liu2018SRUtility} (\citeyear{Liu2018SRUtility}) with a few modifications.

\subsection{L1 and L2 regularization}

We employed L1 and L2 regularization in two different ways: on the weights of the network or on the activations of the hidden layers.
In both cases, this involves modifying the loss function in Equation (\ref{eq:dqn_loss}) to include a penalty that is a function of the size of the weights or the activations.
For example, for a neural network with one matrix of parameters $\boldsymbol{\theta}$ and no bias term, the L1 and L2-weight-regularized losses are:
\begin{align}
    \label{eq:l1_l2_weights}
    L1_{\text{W}}(\boldsymbol{\theta}) \overset{.}{=} L(\boldsymbol{\theta}) + \lambda 
    \left\| \boldsymbol{\theta}  \right\|_1,  \hspace{50pt}
    L2_{\text{W}}(\boldsymbol{\theta}) \overset{.}{=} L(\boldsymbol{\theta}) + \lambda
    \left\| \boldsymbol{\theta}  \right\|^2_2, 
\end{align}
where $L(\boldsymbol{\theta})$ is defined as in Equation (\ref{eq:dqn_loss}), $\lambda \geq 0$, and $\left\| \cdot \right\|_1$ and $\left\| \cdot \right\|^2_2$ correspond to the L1 and L2-norm, respectively.
We will refer to these two different type of regularization techniques as $L1_\text{W}$ and $L2_\text{W}$.

In the case of the L1 and L2 regularization on the activations, consider a neural network with input $\boldsymbol{x}$, weights $\boldsymbol{\theta}$, no bias term, and activation function $g$. 
In such a case, the activations of the hidden layer are computed as $\boldsymbol{y} = g(\boldsymbol{\theta}^\top \boldsymbol{x})$, where $g$ is applied component-wise. 
In this case, we define the L1 and L2-regularized losses as:
\begin{align}
    \label{eq:l1_l2_activations}
    L1_{\text{A}}(\boldsymbol{\theta}) \overset{.}{=} L(\boldsymbol{\theta}) + \lambda 
    \left\| \boldsymbol{y}  \right\|_1,  \hspace{50pt}
    L2_{\text{A}}(\boldsymbol{\theta}) \overset{.}{=} L(\boldsymbol{\theta}) + \lambda
    \left\| \boldsymbol{y}  \right\|^2_2, 
\end{align}
where everything is the same as in the previous equations except for the norm which is applied to the activations $\boldsymbol{y}$.
We will refer to these two regularization techniques as $L1_{\text{A}}$ and $L2_{\text{A}}$. 

\subsection{Distributional Regularizers}

An alternative to norm-based regularizers are distributional regularizers,
which were introduced by \citeauthor{Nguyen2011Sparse} (\citeyear{Nguyen2011Sparse}) and then further developed by \citeauthor{Liu2018SRUtility} (\citeyear{Liu2018SRUtility}).
In this section, we propose another way to use this regularization method with a different type  of distribution.

The main idea of this type of regularization is to model the activations of the neurons of each layer after a target exponential family distribution with natural parameter $\beta$ that specifies the level of sparsity of the layer (e.g., $y_{k,i} \sim p_\beta$ for layer $k$, neuron $i$, and an exponential family distribution $p_\beta$).
To encourage this, a regularization penalty is added according to how far the empirical distribution of the activation of a neuron, $p_{\hat{\beta}}$, is from the target distribution.
The regularization penalty is proportional to the KL-divergence between the two distributions, $KL(p_\beta \| p_{\hat{\beta}})$. 
However, since it is very difficult for the empirical distribution to exactly match the target distribution, \citeauthor{Liu2018SRUtility} (\citeyear{Liu2018SRUtility}) relaxed this condition by comparing the distance of the empirical distribution to a set of target distributions, e.g., $Q_B \overset{.}{=} \{ p_\beta | \beta \in B = [\beta_1, \beta_2], 0 \leq \beta_1 < \beta2\}$, and defined such a distance as the Set KL-divergence $SKL(Q_B \| p_{\hat{\beta}}) \overset{.}{=} \min_p KL(p \| p_{\hat{\beta}})$.
They showed that, if $B$ is a convex set, then the SKL-divergence has the form:
\begin{equation}
  SKL(Q_B \| p_{\hat{\beta}}) =
    \begin{cases}
      KL(p_{\beta_2} \| p_{\hat{\beta}}), & \text{if} \ \ \hat{\beta} > \beta_2  \\
      KL(p_{\beta_1} \| p_{\hat{\beta}}), & \text{if} \ \ \hat{\beta} < \beta_1   \\
      0, & \text{otherwise}
    \end{cases}       
\end{equation}
We can then add this regularization term weighted by a positive regularization factor $\lambda_{KL}$ to the DQN loss in Equation (\ref{eq:dqn_loss}) to induce a sparsity level between $\beta_2$ and $\beta_1$.
The loss function is well defined since $\hat{\beta}$ can be estimated for each neuron from the mini-batch sampled from the experience replay buffer and is differentiable with respect to the parameters of the network, $\boldsymbol{\theta}_t$. 
If we model the activations of the neurons as an Exponential distribution and use the convex set $B=(0,\beta]$, then the set KL-divergence is:
%
\begin{equation}
  SKL(Q_B \| p_{\hat{\beta}}) =
    \begin{cases}
      \text{log}\hat{\beta} + \frac{\beta}{\hat{\beta}} - \text{log}\beta - 1, & \text{if} \ \ \hat{\beta} > \beta \\
      0, & \text{otherwise}
    \end{cases}
\end{equation}
We refer to this method as $DR_e$. 
We also test a another type of distributional regularizer where instead of modeling each individual neuron as an Exponential distribution, we model each layer as a Gamma distribution with natural parameter $\beta$ and shape parameter $\alpha$ equal to the size of the layer $n$. 
This should encourage the entire layer to have an average activation between $0$ and $\beta$
, but not enforce a specific level of sparsity for each individual neuron. 
The SKL-divergence is the same as for $DR_e$ but multiplied by $n$, and $\hat{\beta}$ can be estimated by averaging all the activations in the layer.
We refer to this regularization method as $DR_g$.

\subsection{Dropout}

We also study the effect of Dropout \parencite{Hinton2012Dropout1} on the representation learned by DQN. 
In this type of regularization, a random number of units is dropped from a layer with certain probability.
In practice, this means that each neuron is set to zero with a probability of $p$ for each mini-batch of data during training.
During evaluation, all the neurons are active and weighted by $p$, which is equivalent to using the average activation of the corresponding neuron.
We distinguish between the two different ways to process the data as training and evaluation.
In our DQN architecture, the target network, $\boldsymbol{\theta}^-_t$ in Equation (\ref{eq:dqn_loss}), is always set to evaluation.
On the other hand, the policy network, $\boldsymbol{\theta}_t$ in Equation (\ref{eq:dqn_loss}), is set to evaluation when choosing actions and is set to training when computing a training step.

\section{Experiments}

Our goal is to investigate whether it is possible to learn a sparse representation incrementally and whether there is a benefit from doing so.
To accomplish this goal, we studied three main hypotheses that were formulated based on the work by \citeauthor{Liu2018SRUtility} (\citeyear{Liu2018SRUtility}):
\begin{enumerate}
    \item $L1_\text{W}$ and $L2_\text{W}$ will learn a denser representation than DQN, whereas $L1_\text{A}$, $L2_\text{A}$, Dropout, $DR_e$, and $DR_g$ will learn a sparser representation than DQN.
    \item Methods that learned a sparse representation will perform better than methods that learned a dense representation.
    \item The performance of the methods that learned a sparse representation will be more robust to the size of the experience replay buffer than the performance of methods that learned a dense representation.
\end{enumerate}
To test these hypotheses, we used the benchmark domains mountain car and 4-dimensional catcher.
We trained each agent for 200k steps in mountain car and 500k steps in catcher without resets.
The measure of performance was the cumulative reward over the whole training period.
For this reason, we modified the mountain car environment by giving a reward of 0, instead of -1, when the agent reaches the terminal state; this way, the cumulative reward is informative of the learning progress.
We chose these environments so that our results can be directly compared to the results from \citeauthor{Liu2018SRUtility} (\citeyear{Liu2018SRUtility}) and because they are light enough to allow for a large number of runs, which allows us to make statistical arguments about the performance of each algorithm. 

We used the same architecture in all of our experiments consisting of two hidden layers with 32 and 256 units, respectively, ReLU activations, and a linear output layer with no bias term. 
We initialized the weights of each layer of size $n$ in the network according to a zero-mean Gaussian distribution and variance of $2/n$; the bias terms of the hidden layers are all initialized to zero \parencite{He2015Init}.
To minimize the loss function, we used the Adam optimizer \parencite{King2015Adam} with $\beta_1=0.9$, $\beta_2=0.999$ and $\epsilon=1 \times 10^{-8}$.
The mini-batch size was set to 32 for all the experiments.

For each different method, we found the parameter combination that maximized the cumulative reward by performing a grid search over the learning rate and the method's parameters using 30 samples for each parameter combination.
For DQN, we tested buffer size values in \{100, 1k, 5k, 20k, 80k\} and target network update frequencies in \{10, 50, 100, 200, 400\}.
All the other methods used the same buffer size and target network as the best combination found for DQN. 
For more details about the values of each parameter used in the grid search, see Appendix A.
Finally, in the case of $L1_\text{W}$, $L2_\text{W}$, and dropout, regularization was applied to all the parameters, or all the activations in the case of dropout, of the representation $\boldsymbol{\phi}_{\boldsymbol{\theta}}$. 
On the other hand, for $L1_\text{A}$, $L2_\text{A}$, $DR_e$, and $DR_g$, regularization was applied only to the activations of the last layer of the representation. 
This was done to emulate the experimental setup of \citeauthor{Liu2018SRUtility} (\citeyear{Liu2018SRUtility}).

\subsection{Hypothesis 1: Learning Sparse Representations}
To test our first hypothesis, we first found the best combination of buffer size and target network update frequency for a DQN agent (5k and 10, respectively, for mountain car and 80k and 400, respectively, for catcher).
Then, we fixed the buffer size and target network update frequency to be the same as for DQN and swept over each of the parameters of each regularization method to find the best parameter combination.
After finding the best parameter combination for each different method, we ran another 500 runs to eliminate possible maximization bias.
Our analyses were performed on the second hidden layer of the network at the end of training.

\begin{table}[t]
  \caption{\textbf{(A)} Measures of sparsity: activation overlap (Overlap), live neurons (Neurons), and normalized activation overlap (Normalized Overlap).
  \textbf{(B)} Performance: cumulative reward over the entire training period (200k steps for Mountain car and 500k for Catcher).
  The sample average (Avg) and margin of error of the 95\% confidence interval (ME) were computed based on 500 independent runs.
  }
  \label{tbl:activation_overlap}
  \centering
  \begin{tabular}{lccccccccc}
    \toprule
    & \multicolumn{6}{c}{\textbf{(A) Measures of Sparsity}} & \multicolumn{2}{c}{\textbf{(B) Performance}} \\
    \cmidrule(r){2-7}
    \cmidrule(r){8-9}
    & \multicolumn{2}{c}{Overlap} & \multicolumn{2}{c}{Neurons} & \multicolumn{2}{c}{Normalized Overlap} & \multicolumn{2}{c}{Cumulative Reward} \\
    \cmidrule(r){2-3}
    \cmidrule(r){4-5}
    \cmidrule(r){6-7}
    \cmidrule(r){8-9}
    Method          & Avg       & ME        & Avg       & ME        & Avg       & ME        & Avg           
                    & ME        \\
    \midrule
    & \multicolumn{8}{c}{\textbf{Mountain Car}} \\
    \midrule
    DQN             & 17.92     & 0.64      & 29.21     & 1.14      & 0.64      & 0.01      & -198 884.57   
                    & 12.61     \\
    Dropout         & 85.8      & 2.01      & 164.35    & 1.08      & 0.53      & 0.014     & -198 970.4
                    & 14.27     \\
    $DR_e$          & 13.26     & 0.47      & 21.04     & 0.86      & 0.65      & 0.01      & -198 869.49
                    & 12.28     \\
    $DR_g$          & 12.96     & 0.55      & 21.05     & 0.91      & 0.63      & 0.01      & -198 870.35
                    & 11.92     \\
    $L1_\text{A}$   & 11.2      & 0.43      & 18.25     & 0.74      & 0.63      & 0.011     & -198 872.44
                    & 10.09     \\
    $L1_\text{W}$   & 93.66     & 1.94      & 207.75    & 1.54      & 0.45      & 0.007     & \textbf{-198 593.53}
                    & \textbf{8.25}      \\
    $L2_\text{A}$   & 4.52      & 0.11      & 22.21     & 0.73      & 0.22      & 0.005     & \textbf{-198 598.9}
                    & \textbf{4.48}      \\
    $L2_\text{W}$   & 39.52     & 0.73      & 116.92    & 1.25      & 0.34      & 0.005     & -198 633.16
                    & 6.04      \\
    \midrule
    & \multicolumn{8}{c}{\textbf{Catcher}} \\
    \midrule
    DQN             & 58.42     & 0.81      & 154.51    & 1.09      & 0.38      & 0.004     & 11 657.88 
    & 42.23     \\
    Dropout         & 101.26    & 0.99      & 243.33    & 0.48      & 0.42      & 0.004     & 9 565.69 
    & 97.22     \\
    $DR_e$          & 53.7      & 0.92      & 158.48    & 1.15      & 0.34      & 0.005     & 11 730.06  
    & 41.05     \\
    $DR_g$          & 49.17     & 0.6       & 197.97    & 0.9       & 0.25      & 0.003     & \textbf{11 868.85}
    & \textbf{69.21}     \\
    $L1_\text{A}$   & 5.9       & 0.45      & 37.86     & 0.84      & 0.15      & 0.008     & 10 666.23  
    & 114.84    \\
    $L1_\text{W}$   & 90.28     & 1.6       & 161.39    & 1.77      & 0.56      & 0.008     & 11 370.72  
    & 99.72     \\
    $L2_\text{A}$   & 35.94     & 0.94      & 118.66    & 1.35      & 0.3       & 0.006     & \textbf{11 874.5}   
    & \textbf{68.22}     \\
    $L2_\text{W}$   & 72.34     & 0.81      & 182.43    & 1.37      & 0.4       & 0.004     & 11 746.97  
    & 44.56     \\
    \bottomrule
  \end{tabular}
  \vspace{-3mm}
\end{table}

To study the sparsity of the learned representation, we computed the version of activation overlap proposed by \citeauthor{Liu2018SRUtility} (\citeyear{Liu2018SRUtility}).
For two observations $\boldsymbol{x}_1$ and $\boldsymbol{x}_2$ and a hidden layer with $n$ neurons, i.e., $\{ y_i \}_{1 \leq i \leq n}$, the activation overlap is:
\begin{equation}
    \label{eq:activation_overlap}
    \sum^n_{i=1} \mathbb{1} [ (y_i(\boldsymbol{x}_1) > 0)
    \wedge (y_i(\boldsymbol{x}_2) > 0) ].
\end{equation}
To compute this measure, we covered the state space with a grid with 10k vertices by partitioning each dimension in the mountain car environment into 100 equal partitions and each dimension in the catcher environment into 10 equal partitions.
We computed the activation overlap on each pair of vertices in the grid and averaged over 500 runs.
As we were computing the activation overlap we found that many methods had a large number of dead neurons (neurons that were zero for every observation in the data set) and noticed that the measure in Equation (\ref{eq:activation_overlap}) did not capture this. 
Consequently, a method can appear to have low activation overlap because it retained a small number of live neurons.
In Table \ref{tbl:activation_overlap}A, we present the average activation overlap, the number of live neurons, and the normalized activation overlap---normalized by the number of live neurons---along with the margin of error of the 95\% confidence interval.

In both environments, we found that a higher activation overlap corresponded to a higher number of live neurons.
On the other hand, the normalized activation overlap did not show any correspondence to the number of live neurons. 
This is problematic since depending on the measure, we can draw different conclusions about the sparsity of the learned representation of each algorithm, which raises the question: what measure of overlap should we use?

\begin{figure}[t]
  \centering
  \includegraphics[width=0.9\textwidth]{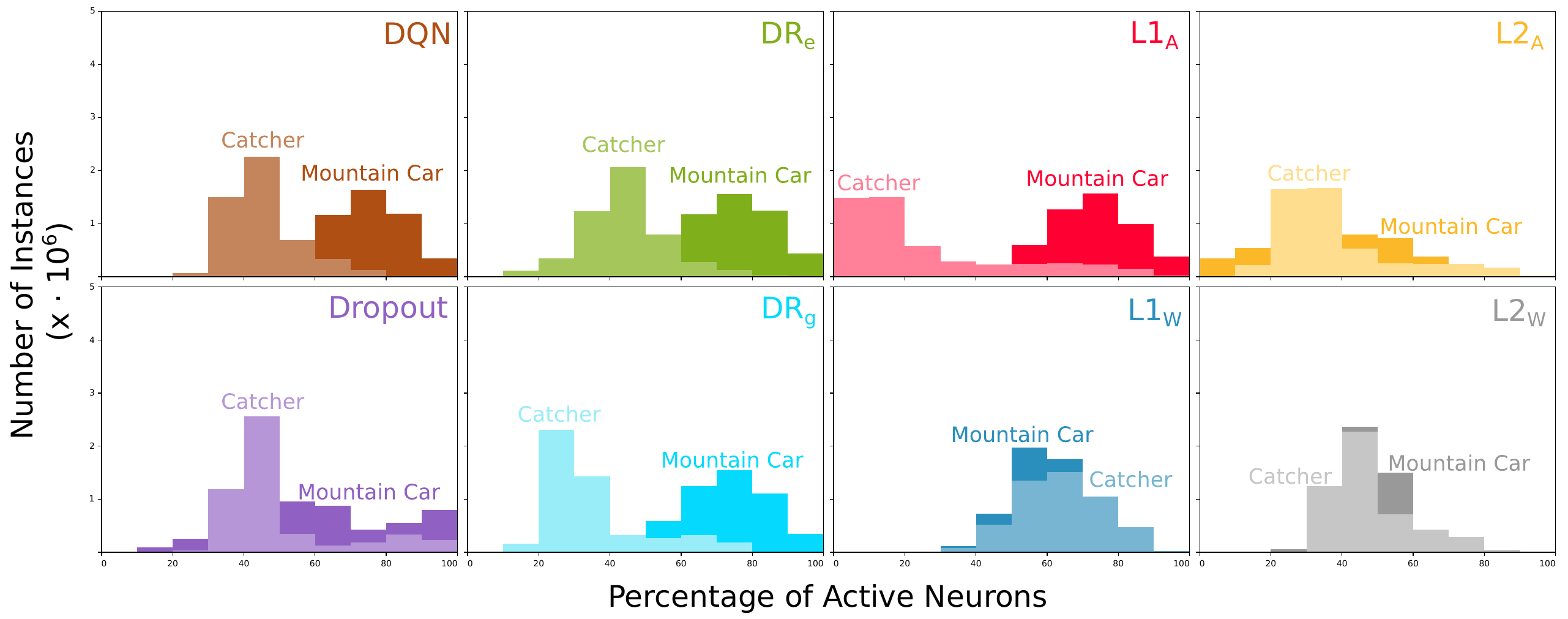}
  \caption{Instance sparsity for DQN and each of the different regularization methods accumulated over 500 runs, excluding dead neurons.
  \vspace{-5mm}}
  \label{fig:instance_sparsity}
\end{figure}

To corroborate the results in Table \ref{tbl:activation_overlap}A, we computed the instance sparsity measure \parencite{Liu2018SRUtility} for each different method using the same samples used to compute the activation overlap.
The instance sparsity corresponds to the percentage of active neurons (excluding dead neurons) for each instance in a data set.
A sparse representation should result in small percentage of active neurons for each instance. 
Figure \ref{fig:instance_sparsity} shows the instance sparsity of each different method for each different environment aggregated over 500 runs; we used light colours for catcher and dark colours for mountain car.

The results show that there is not a clear relationship between the activation overlap and the instance sparsity measures. 
For example, in mountain car, both $L2_\text{W}$ and $L2_\text{A}$ resulted in higher activation overlap than DQN, which indicates that both of this methods learned a denser representation than DQN if we accept activation overlap as a measure of sparsity.
However, the instance sparsity plot shows that the representation learned by $L2_\text{W}$ and $L2_\text{A}$ is sparser than the representation learned by DQN, contradicting the conclusion drawn from the activation overlap.
On the other hand, the normalized activation overlap shows a strong relationship with the instance sparsity plots.
The clearest example is $L1_\text{A}$, which shows a similar level of sparsity as DQN in mountain car, but a higher level of sparsity than DQN in catcher according to the instance sparsity plots.
The normalized activation overlap corroborates this conclusion, unlike the activation overlap without normalization.
Consequently, we will use the normalized activation overlap as the main measure of sparsity.

Overall, the results show that it is possible to learn a sparse representation incrementally by using appropriate regularization. 
However, $L1_\text{W}$ and $L2_\text{W}$ do not necessarily result in a denser representation than DQN.
Moreover, Dropout, $DR_e$, $DR_g$, and $L1_\text{A}$ do not consistently result in a sparser representation than DQN.
The only method that resulted in a sparser representation than DQN in both environments was $L2_\text{A}$.
Since it seems difficult to learn a sparse representation incrementally, one must ask: is there any benefit from learning sparse representations?
\vspace{-1mm}
%
\subsection{Hypothesis 2: The utility of sparse representations}
To test hypothesis 2, we took a closer look at the performance of the algorithms from the previous experiment.
If we accept hypothesis 2 to be true, then we would expect the methods with a smaller normalized activation overlap to have the best performance among all the algorithms.
In other words, we would expect $L2_\text{A}$ to perform the best in mountain car, and $L1_\text{A}$ to perform the best in catcher. 
Table \ref{tbl:activation_overlap}B shows that this is true in mountain car, where $L2_\text{A}$ and $L1_\text{W}$ had the best performance.
However, we can already see evidence of a more complex effect.
For instance, $L1_\text{W}$ learned a denser representation than $L2_\text{W}$ in mountain car, yet it resulted in better performance.
Similarly, in catcher, $L1_\text{A}$---the method with the lowest normalized activation overlap---performed worse than many of the methods that resulted in denser representations. 

The results indicate that learning a sparse representation can improve performance, but only if this does not result in a large number of dead neurons.
Conversely, learning a slightly denser representation, as in the case of $L1_\text{W}$ compared to $L2_\text{W}$ in mountain car, can result in good performance as long as many neurons stay alive. 
This suggests that methods that learn a sparse representation while preserving as many live neurons as possible would perform better than methods that solely learn a sparse representation or solely preserve as many live neurons as possible. 
We postpone the investigation of this hypothesis for future work.
\begin{figure}[t]
  \centering
  \includegraphics[width=1\textwidth]{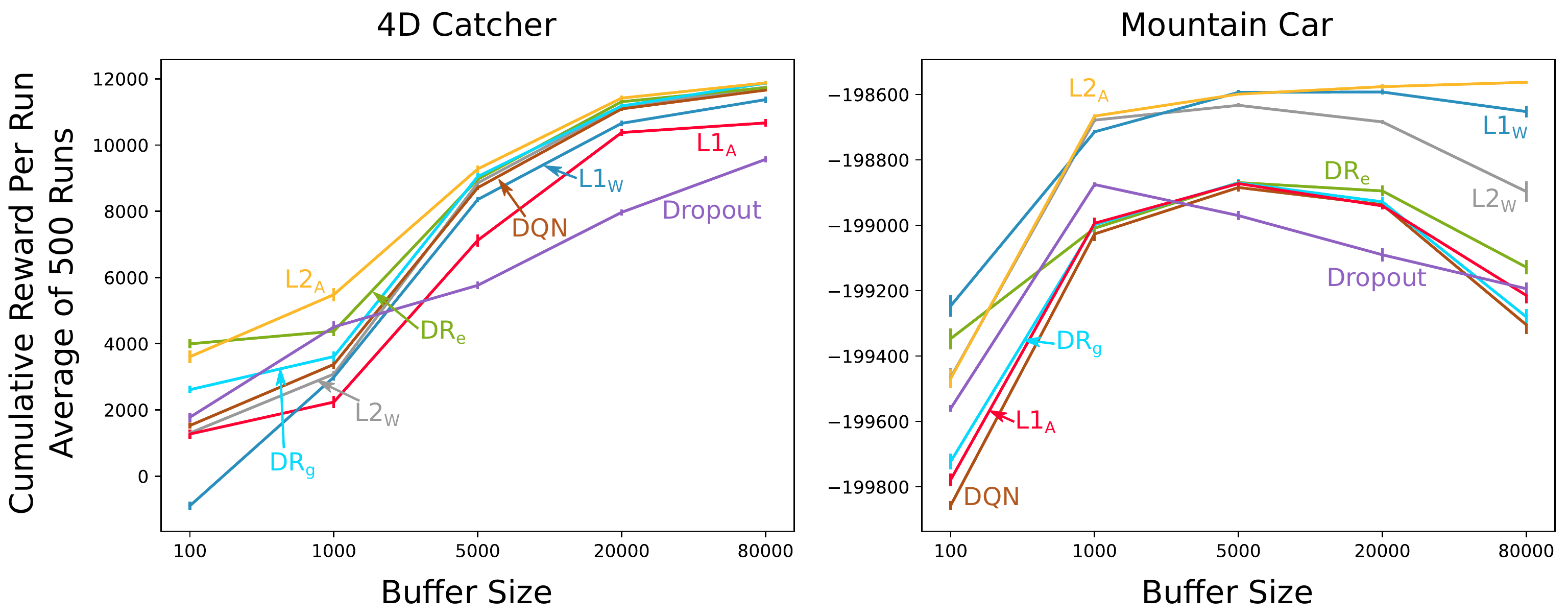}
  \caption{Cumulative reward over the entire training period for each method with different buffer size values for both environments.
  The results are averaged over 500 independent runs and the shaded regions correspond to a 95\% confidence interval.
  \vspace{-2.5mm}}
  \label{fig:perf_measure}
\end{figure}
%
\subsection{Hypothesis 3: Robustness of Sparse Representations to the Replay Buffer Size}
Beyond improving performance in terms of cumulative reward, sparse representations may also be useful for overcoming the catastrophic interference problem often encountered in DNNs.
Since the experience replay buffer mitigates the catastrophic interference suffered by a DNN, it should be possible to control the amount of interference by adjusting the size of the buffer.
In fact, previous results have shown that either a buffer too small or too big can have a negative effect in performance \parencite{Zhang2017ER, LiuR2017Effects} suggesting the occurrence of catastrophic interference at either extreme.
Consequently, if sparse representations help mitigate catastrophic interference, we would expect that the performance of those methods that learned a sparser representation to be more robust to the size of the experience replay buffer.

To test this hypothesis, we implemented several agents with buffer size values of 100, 1k, 2k, 5k, 20k, and 80k.
For each of these values, we found the best parameter combination for DQN and each of the different regularization methods.
The regularization methods used the same target network update frequency as DQN to eliminate possible confounding effects.
After finding the best parameter combination for each different method, we ran each method for another 500 runs to eliminate possible maximization bias.

Our results---Figure \ref{fig:perf_measure}B---provide evidence in favour of our hypothesis.
Methods that learn a sparser representation were more robust to the size of the experience replay buffer.
This is most evident in mountain car where the performances of $L1_\text{W}$ and $L2_\text{A}$, the two methods with the lowest normalized activation overlap, are more robust to the effect of the buffer size.
A similar effect can be observed in catcher to a lesser degree; the performance of $L2_\text{A}$ is more robust to the effect of the buffer size.
However, once again we found evidence of a more complex effect.
If learning a sparse representation was solely responsible for the robustness of each method to the size of the experience replay buffer, then we would expect $L1_\text{A}$ to be more robust to the effect of the buffer size in catcher, yet its performance is one of the worse among all the methods.
We hypothesize that this effect is the result of $L1_\text{A}$ regularization killing too many neurons during learning.
%
\section{Conclusions}
In this paper we empirically demonstrated that it is possible to learn a sparse representation and the action-value function simultaneously. 
Moreover, we corroborated the results from \citeauthor{Liu2018SRUtility} (\citeyear{Liu2018SRUtility}) by showing that sparse representations are useful for improving performance and for overcoming catastrophic interference in reinforcement learning. 
Most importantly, we found that how we learn is just as important as what we learn; learning a sparse representation seems to be useful for improving performance, but killing too many neurons in the process could be counterproductive. 
This insight suggests that we should strive for methods that learn a sparse representation while retaining as many live neurons as possible; however, further work is needed to confirm this hypothesis. 

\printbibliography

\newpage
\section*{Appendix A: Grid Search}
In order to find the best parameters in our experiments we performed a grid search with a sample size of 30.
To evaluate each parameter combination we compared the 95\% confidence interval of the cumulative reward over the whole training period and selected the parameter combination that resulted in the highest lower confidence bound. 
This criteria selects methods that achieve the highest cumulative reward and also has a small variance. 
Once we found the best parameter combination, we reran every method for another 500 runs in order to eliminate maximization bias. 

These are the values of the parameters that we used in our grid search:
\begin{table}[h]
    \centering
    \begin{tabular}{l|c|l}
         Hyperparameter                 & Method                & Values  \\ \bottomrule \rule{0pt}{1.2\normalbaselineskip}
         \multirow{3}{*}{Learning Rate} & \multirow{3}{*}{All}  & Mountain Car: 0.01, 0.004, 0.001, 0.00025         \\
                                        &                       & Catcher: 0.001, 0.0005, 0.00025, 0.000125,        \\
                                        &                       & 0.0000625, 0.00003125, 0.000015625 
         \\[1mm] \hdashline[3pt/3pt] \rule{0pt}{1.2\normalbaselineskip}
         Experience Replay Buffer Size      & DQN               & 100, 1k, 5k, 20k, 80k
         \\[1mm] \hdashline[3pt/3pt] \rule{0pt}{1.2\normalbaselineskip}
         Target Network Update Frequency    & DQN               & 10, 50, 100, 200, 400
         \\[1mm] \hdashline[3pt/3pt] \rule{0pt}{1.2\normalbaselineskip}
         Dropout Probability ($p$)          & Dropout               & 0.1, 0.2, 0.3, 0.4, 0.5
         \\[1mm] \hdashline[3pt/3pt] \rule{0pt}{1.2\normalbaselineskip}
         Beta Upper Bound ($\beta$)     & $DR_e$, $DR_g$        & 0.1, 0.2, 0.5
         \\[1mm] \hdashline[3pt/3pt] \rule{0pt}{1.2\normalbaselineskip}
         Regularization Factor ($\lambda_\text{KL}$):       & \multirow{2}{*}{$DR_e$, $DR_g$} 
                                                            & \multirow{2}{*}{0.1, 0.01, 0.001} \\
         \hspace{0.5pt} Distributional Regularizers   & & 
         \\[1mm] \hdashline[3pt/3pt] \rule{0pt}{1.2\normalbaselineskip}
         Regularization Factor ($\lambda$):     & $L1_\text{A}$, $L1_\text{W}$, & 0.1, 0.05, 0.01, 0.005, 0.001,  \\
         Normed-Based Regularizers              & $L2_\text{A}$, $L2_\text{W}$  & 0.0005, 0.0001
        \\[1mm] \bottomrule
    \end{tabular}
    \caption{
    Hyper parameters used in the grid search for all of our experiments. Note that we only optimized the experience replay buffer and target network update frequency for DQN. All the regularization methods used the same buffer size and target network update frequency as DQN.
    }
    \label{tab:extended_parameter_table}
\end{table}

\newpage
\section*{Appendix B: Extended Tables}
We omitted the standard deviation in Table \ref{tbl:activation_overlap} because of space and for concreteness.
For similar reasons, in Figure \ref{fig:perf_measure} we did not provide any details about the specific values of the average performance measure and their confidence intervals.
In this appendix we present extended results to facilitate reproducibility and allow the reader to double check our results. 

\begin{table}[h]
  \caption{Measures of sparsity: activation overlap (Overlap), live neurons (Neurons), and normalized activation overlap (Normalized Overlap).
  The sample average (Avg), standard deviation (SD), and margin of error of the 95\% confidence interval (ME) were computed based on 500 independent runs.
  }
  \label{tbl:extended_activation_overlap}
  \centering
  \begin{tabular}{lcccccccccc}
    \toprule
    & \multicolumn{10}{c}{\textbf{Mountain Car}} \\
    \midrule
    & \multicolumn{3}{c}{Overlap} & \multicolumn{3}{c}{Neurons} & \multicolumn{3}{c}{Normalized Overlap} \\
    \cmidrule(r){2-4}
    \cmidrule(r){5-7}
    \cmidrule(r){8-10}
    Method          & Avg       & SD        & ME        & Avg       & SD        & ME        
                    & Avg       & SD        & ME        \\
    \midrule
    DQN             & 17.92     & 7.26      & 0.64      & 29.21     & 12.97     & 1.14      
                    & 0.64      & 0.12      & 0.01      \\
    Dropout         & 85.8      & 22.83     & 2.01      & 164.35    & 12.25     & 1.08
                    & 0.53      & 0.16      & 0.014     \\
    $DR_e$          & 13.26     & 5.36      & 0.47      & 21.04     & 9.77      & 0.86      
                    & 0.65      & 0.11      & 0.01      \\
    $DR_g$          & 12.96     & 6.27      & 0.55      & 21.05     & 10.36     & 0.91      
                    & 0.63      & 0.12      & 0.01      \\
    $L1_\text{A}$   & 11.2      & 4.95      & 0.43      & 18.25     & 8.42      & 0.74      
                    & 0.63      & 0.12      & 0.011     \\
    $L1_\text{W}$   & 93.66     & 22.04     & 1.94      & 207.75    & 17.55     & 1.54      
                    & 0.45      & 0.08      & 0.007     \\
    $L2_\text{A}$   & 4.52      & 1.2       & 0.11      & 22.21     & 8.27      & 0.73      
                    & 0.22      & 0.06      & 0.005     \\
    $L2_\text{W}$   & 39.52     & 8.34      & 0.73      & 116.92    & 14.21     & 1.25      
                    & 0.34      & 0.05      & 0.005     \\
    \midrule
    & \multicolumn{8}{c}{\textbf{Catcher}} \\
    \midrule
    & \multicolumn{3}{c}{Overlap} & \multicolumn{3}{c}{Neurons} & \multicolumn{3}{c}{Normalized Overlap} \\
    \cmidrule(r){2-4}
    \cmidrule(r){5-7}
    \cmidrule(r){8-10}
    Method          & Avg       & SD        & ME        & Avg       & SD        & ME        
                    & Avg       & SD        & ME        \\
    \midrule
    DQN             & 58.42     & 9.17      & 0.81      & 154.51    & 12.35     & 1.09      
                    & 0.38      & 0.05      & 0.004     \\
    Dropout         & 101.26    & 11.3      & 0.99      & 243.33    & 5.52      & 0.48      
                    & 0.42      & 0.05      & 0.004     \\
    $DR_e$          & 53.7      & 10.49     & 0.92      & 158.48    & 13.12     & 1.15      
                    & 0.34      & 0.05      & 0.005     \\
    $DR_g$          & 49.17     & 6.82      & 0.6       & 197.97    & 10.2      & 0.9       
                    & 0.25      & 0.03      & 0.003     \\
    $L1_\text{A}$   & 5.9       & 5.14      & 0.45      & 37.86     & 9.57      & 0.84      
                    & 0.15      & 0.1       & 0.008     \\
    $L1_\text{W}$   & 90.28     & 18.16     & 1.6       & 161.39    & 20.16     & 1.77      
                    & 0.56      & 0.09      & 0.008     \\
    $L2_\text{A}$   & 35.94     & 10.67     & 0.94      & 118.66    & 15.39     & 1.35      
                    & 0.3       & 0.07      & 0.006     \\
    $L2_\text{W}$   & 72.34     & 9.22      & 0.81      & 182.43    & 15.54     & 1.37      
                    & 0.4       & 0.05      & 0.004     \\
    \bottomrule
  \end{tabular}
  \vspace{-3mm}
\end{table}

\begin{table}[h]
  \caption{ Results for all the methods with different buffer sizes in the mountain car environment. The average (Avg), standard deviation (SD), margin of error (ME), and 95\% confidence interval (C.I.) were computed using a sample size of 500.
  }
  \label{tbl:all_results_mc}
  \centering
  \begin{tabular}{lccccc}
    \toprule
    Method                      & Buffer Size   & Avg           & SD        & ME        &   C.I.                        \\
    \midrule
    \multirow{5}{*}{DQN}        & 100           & -199 856.86   & 150.98    & 13.27     & (-199 870.12, -199 843.59)    \\
                                & 1 \text{K}    & -199 026.66   & 244.65    & 21.5      & (-199 048.15, -199 005.16)    \\ 
                                & 5 \text{K}    & \textbf{-198 884.57}   & 143.48    & 12.61     
                                & \textbf{(-198 897.17, -198 871.96)}    \\     
                                & 20 \text{K}   & -198 937.06   & 141.23    & 12.41     & (-198 949.47, -198 924.65)    \\
                                & 80 \text{K}   & -199 304.41   & 314.35    & 27.62     & (-199 332.03, -199 276.79)    \\
    \midrule
    \multirow{5}{*}{DRE}        & 100           & -199 346.84   & 364.26    & 32.01     & (-199 378.84, -199 314.83)    \\
                                & 1 \text{K}    & -199 009.05   & 236.99    & 20.82     & (-199 029.87, -198 988.23)    \\
                                & 5 \text{K}    & \textbf{-198 869.49}   & 139.73    & 12.28     
                                & \textbf{(-198 881.77, -198 857.21)}    \\     
                                & 20 \text{K}   & \textbf{-198 895.09}   & 184.29    & 16.19     & 
                                \textbf{(-198 911.29, -198 878.9)}     \\       
                                & 80 \text{K}   & -199 128.19   & 254.1     & 22.33     & (-199 150.52, -199 105.87)    \\
    \midrule
    \multirow{5}{*}{DRG}        & 100           & -199 722.63   & 271.86    & 23.89     & (-199 746.52, -199 698.74)    \\
                                & 1 \text{K}    & -198 998.54   & 192.22    & 16.89     & (-199 015.43, -198 981.65)    \\
                                & 5 \text{K}    & \textbf{-198 870.35}   & 135.71    & 11.92     
                                & \textbf{(-198 882.27, -198 858.42)}    \\     
                                & 20 \text{K}   & -198 928.31   & 171.58    & 15.08     & (-198 943.39, -198 913.23)    \\
                                & 80 \text{K}   & -199 280.11   & 276.14    & 24.26     & (-199 304.38, -199 255.85)    \\
                                \midrule
    \multirow{5}{*}{L1A}        & 100           & -199 778.29   & 221.7     & 19.48     & (-199 797.77, -199 758.81)    \\
                                & 1 \text{K}    & -198 993.98   & 200.59    & 17.63     & (-199 011.61, -198 976.36)    \\
                                & 5 \text{K}    & \textbf{-198 872.44}   & 114.8     & 10.09     
                                & \textbf{(-198 882.52, -198 862.35)}    \\     
                                & 20 \text{K}   & -198 940.82   & 124.59    & 10.95     & (-198 951.77, -198 929.87)    \\
                                & 80 \text{K}   & -199 214.54   & 278.61    & 24.48     & (-199 239.02, -199 190.06)    \\
                                \midrule
    \multirow{5}{*}{L1W}        & 100           & -199 246.67   & 376.84    & 33.11     & (-199 279.79, -199 213.56)    \\
                                & 1 \text{K}    & -198 714.28   & 57.01     & 5.01      & (-198 719.29, -198 709.27)    \\
                                & 5 \text{K}    & \textbf{-198 593.53}   & 93.91     & 8.25      
                                & \textbf{(-198 601.78, -198 585.28)}    \\     
                                & 20 \text{K}   & \textbf{-198 592.51}   & 94.1      & 8.27      
                                & \textbf{(-198 600.78, -198 584.24)}    \\     
                                & 80 \text{K}   & -198 652.34   & 200.2     & 17.59     & (-198 669.93, -198 634.75)    \\
                                \midrule
    \multirow{5}{*}{L2A}        & 100           & -199 468.76   & 328.71    & 28.88     & (-199 497.64, -199 439.88)    \\
                                & 1 \text{K}    & -198 666.12   & 58.59     & 5.15      & (-198 671.26, -198 660.97)    \\
                                & 5 \text{K}    & -198 598.9    & 50.96     & 4.48      & (-198 603.38, -198 594.42)    \\
                                & 20 \text{K}   & \textbf{-198 576.14}   & 74.58     & 6.55      
                                & \textbf{(-198 582.69, -198 569.59)}    \\     
                                & 80 \text{K}   & \textbf{-198 562.71}    & 55.41     & 4.87      
                                & \textbf{(-198 567.58, -198 557.84)}    \\     
                                \midrule
    \multirow{5}{*}{L2W}        & 100           & -199 465.72   & 339.12    & 29.8      & (-199495.52, -199435.92)      \\
                                & 1 \text{K}    & -198 678.4    & 75.54     & 6.64      & (-198 685.04, -198 671.76)    \\
                                & 5 \text{K}    & \textbf{-198 633.16}   & 68.7      & 6.04      
                                & \textbf{(-198 639.2, -198 627.13)}     \\     
                                & 20 \text{K}   & -198 683.92   & 73.24     & 6.44      & (-198 690.35, -198 677.48)    \\
                                & 80 \text{K}   & -198 896.84   & 356.71    & 31.34     & (-198 928.19, -198 865.5)     \\
                                \midrule
    \multirow{5}{*}{Dropout}    & 100           & -199 560.45   & 115.59    & 10.16     & (-199 570.6, -199 550.29)     \\
                                & 1 \text{K}    & \textbf{-198 875.45}   & 80.23     & 7.05      
                                & \textbf{(-198 882.5, -198 868.4)}      \\
                                & 5 \text{K}    & -198 970.4    & 162.35    & 14.27     & (-198 984.66, -198 956.13)    \\
                                & 20 \text{K}   & -199 090.34   & 232.55    & 20.43     & (-199 110.78, -199 069.91)    \\
                                & 80 \text{K}   & -199 194.09   & 211.28    & 18.56     & (-199 212.65, -199 175.52)    \\
    \bottomrule
  \end{tabular}
\end{table}

\begin{table}[h]
  \caption{ Results for all the methods with different buffer sizes in the catcher environment. The average (Avg), standard deviation (SD), margin of error (ME), and 95\% confidence interval (C.I.) were computed using a sample size of 500.
  }
  \label{tbl:all_results_catcher}
  \centering
  \begin{tabular}{lccccc}
    \toprule
    Method                      & Buffer Size   & Avg           & SD        & ME        &   C.I.                \\
    \midrule
    \multirow{5}{*}{DQN}        & 100           & 1529.36       & 992.14    & 87.17     & (1442.19, 1616.54)    \\
                                & 1 \text{K}    & 3374.82       & 1606.29   & 141.14    & (3233.68, 3515.96)    \\
                                & 5 \text{K}    & 8711.13       & 813.98    & 71.52     & (8639.61, 8782.65)    \\
                                & 20 \text{K}   & 11090.68      & 638.03    & 56.06     & (11034.62, 11146.74)  \\
                                & 80 \text{K}   & \textbf{11657.88}      & 480.58    & 42.23     
                                & \textbf{(11615.65, 11700.11)}  \\
    \midrule
    \multirow{5}{*}{DRE}        & 100           & 3995.81       & 1522.67   & 133.79    & (3862.02, 4129.6)     \\
                                & 1 \text{K}    & 4377.34       & 1704.83   & 149.8     & (4227.54, 4527.14)    \\
                                & 5 \text{K}    & 8968.46       & 1247.77   & 109.64    & (8858.82, 9078.1)     \\
                                & 20 \text{K}   & 11308.25      & 612.84    & 53.85     & (11254.4, 11362.1)    \\
                                & 80 \text{K}   & \textbf{11730.06}      & 467.21    & 41.05     
                                & \textbf{(11689.01, 11771.12)}  \\
    \midrule
    \multirow{5}{*}{DRG}        & 100           & 2610.76       & 1316.8    & 115.7     & (2495.06, 2726.47)    \\
                                & 1 \text{K}    & 3612.21       & 1819.77   & 159.89    & (3452.32, 3772.11)    \\
                                & 5 \text{K}    & 9047.32       & 1174.89   & 103.23    & (8944.09, 9150.56)    \\
                                & 20 \text{K}   & 11178.37      & 576.01    & 50.61     & (11127.76, 11228.98)  \\
                                & 80 \text{K}   & \textbf{11868.85}      & 787.65    & 69.21     
                                & \textbf{(11799.64, 11938.06)}  \\
                                \midrule
    \multirow{5}{*}{L1A}        & 100           & 1273.31       & 1597.59   & 140.37    & (1132.94, 1413.68)    \\
                                & 1 \text{K}    & 2237.3        & 2092.8    & 183.89    & (2053.41, 2421.18)    \\
                                & 5 \text{K}    & 7121.2        & 2104.6    & 184.92    & (6936.28, 7306.12)    \\
                                & 20 \text{K}   & 10376.92      & 1225.84   & 107.71    & (10269.21, 10484.63)  \\
                                & 80 \text{K}   & \textbf{10666.23}      & 1307.04   & 114.84    
                                & \textbf{(10551.39, 10781.08)}  \\
                                \midrule
    \multirow{5}{*}{L1W}        & 100           & -894.34       & 1368.01   & 120.2     & (-1014.54, -774.14)   \\
                                & 1 \text{K}    & 2989.15       & 1163.98   & 102.27    & (2886.87, 3091.42)    \\
                                & 5 \text{K}    & 8353.5        & 763.06    & 67.05     & (8286.45, 8420.54)    \\
                                & 20 \text{K}   & 10655.52      & 910.01    & 79.96     & (10575.56, 10735.47)  \\
                                & 80 \text{K}   & \textbf{11370.72}      & 1134.97   & 99.72     
                                & \textbf{(11270.99, 11470.44)}  \\
                                \midrule
    \multirow{5}{*}{L2A}        & 100           & 3606.2        & 2138.51   & 187.9     & (3418.3, 3794.1)      \\
                                & 1 \text{K}    & 5481.86       & 2317.08   & 203.59    & (5278.27, 5685.45)    \\
                                & 5 \text{K}    & 9276.56       & 1209.6    & 106.28    & (9170.28, 9382.84)    \\
                                & 20 \text{K}   & 11417.48      & 832.52    & 73.15     & (11344.33, 11490.63)  \\
                                & 80 \text{K}   & \textbf{11874.5}       & 776.4     & 68.22     
                                & \textbf{(11806.29, 11942.72)}  \\
                                \midrule
    \multirow{5}{*}{L2W}        & 100           & 1302.85       & 987.5     & 86.77     & (1216.09, 1389.62)    \\
                                & 1 \text{K}    & 3082.81       & 1098.62   & 96.53     & (2986.28, 3179.34)    \\
                                & 5 \text{K}    & 8853.96       & 766.19    & 67.32     & (8786.63, 8921.28)    \\
                                & 20 \text{K}   & 11167.58      & 580.9     & 51.04     & (11116.54, 11218.62)  \\
                                & 80 \text{K}   & \textbf{11746.97}      & 507.19    & 44.56     
                                & \textbf{(11702.41, 11791.53)}  \\
                                \midrule
    \multirow{5}{*}{Dropout}    & 100           & 1779.72       & 1580.31   & 138.85    & (1640.86, 1918.57)    \\
                                & 1 \text{K}    & 4505.58       & 1891.71   & 166.22    & (4339.36, 4671.79)    \\
                                & 5 \text{K}    & 5762.51       & 1299.27   & 114.16    & (5648.35, 5876.67)    \\
                                & 20 \text{K}   & 7967.46       & 1058.12   & 92.97     & (7874.49, 8060.43)    \\
                                & 80 \text{K}   & \textbf{9565.69}       & 1106.42   & 97.22     
                                & \textbf{(9468.47, 9662.9)}     \\
    \bottomrule
  \end{tabular}
\end{table}

\end{document}